\documentclass[10pt,twocolumn,letterpaper]{article}

\usepackage{iccv}
\usepackage{times}
\usepackage{epsfig}
\usepackage{graphicx}
\usepackage{amsmath}
\usepackage{amssymb}

\usepackage{subfiles}
\usepackage{multirow}
\usepackage{tabularx}
\usepackage{float}
\usepackage{stfloats}
\usepackage{diagbox}
\usepackage{indentfirst}
\usepackage{enumitem}
\usepackage{algorithmicx}
\usepackage{amsmath}
\usepackage{pifont}
\usepackage{booktabs}
\usepackage{bbm}
\usepackage[ruled,vlined,linesnumbered]{algorithm2e}
\DeclareMathAlphabet\mathzapf       {T1}{pzc} {mb} {it}

\newcommand{\veryshortarrow}[1][6pt]{\mathrel{%
   \hbox{\rule[\dimexpr\fontdimen22\textfont2-.2pt\relax]{#1}{.4pt}}%
   \mkern-4mu\hbox{\usefont{U}{lasy}{m}{n}\symbol{41}}}}

\usepackage[pagebackref=true,breaklinks=true,letterpaper=true,colorlinks,bookmarks=false]{hyperref}

\iccvfinalcopy 

\def\iccvPaperID{xxxx} 

\ificcvfinal\fi

\begin{document}

\title{Differentiable Network Adaption with Elastic Search Space}

\author{Shaopeng Guo \quad Yujie Wang  \quad Kun Yuan \quad Quanquan Li \\
SenseTime Research\\
{\tt\small \{guoshaopeng, wangyujie, yuankun, liquanquan\}@sensetime.com}}

\maketitle
\ificcvfinal\thispagestyle{empty}\fi

\begin{abstract}
In this paper we propose a novel network adaption method called Differentiable Network Adaption (DNA), which can adapt an existing network to a specific computation budget by adjusting the width and depth in a differentiable manner. The gradient-based optimization allows DNA to achieve an automatic optimization of width and depth rather than previous heuristic methods that heavily rely on human priors. Moreover, we propose a new elastic search space that can flexibly condense or expand during the optimization process, allowing the network optimization of width and depth in a bi-direction manner. By DNA, we successfully achieve network architecture optimization by condensing and expanding in both width and depth dimensions. Extensive experiments on ImageNet demonstrate that DNA can adapt the existing network to meet different targeted computation requirements with better performance than previous methods. What's more, DNA can further improve the performance of high-accuracy networks obtained by state-of-the-art neural architecture search methods such as EfficientNet and MobileNet-v3.

\end{abstract}
\section{Introduction}
The design of neural network architectures ~\cite{krizhevsky2012imagenet,szegedy2015going,simonyan2014very,DBLP:journals/corr/HeZRS15, DBLP:journals/corr/HowardZCKWWAA17,DBLP:journals/corr/abs-1801-04381} has been lead to significant accuracy improvements.
In real-world applications on mobiles or other embedded devices, it is usually necessary to adapt an existing network to a larger or smaller computation budget to squeeze its performance on different devices.
The computation overhead of the network is closely related to its depth and width, i.e. the number of layers in the network and the number of channels in each layer.
However, it requires substantial effort to find the optimal solution manually.

Recent works have been proposed trying to maximize accuracy while satisfying the computation constraints by optimizing the width or depth of networks.
Many of them only consider \textit{single-direction optimization} for existing networks, which only either condense~\cite{yang2018netadapt,DBLP:conf/cvpr/GuoWLY20,DBLP:journals/corr/abs-1905-09717,DBLP:journals/corr/abs-1804-03230} or expand~\cite{DBLP:journals/corr/abs-1906-02909} the existing networks.
Network adaption, on the other hand, allows \textit{bi-direction optimization}, which can both condense and expand in the dimension of width and depth.
However, previous network adaption works still suffer several limitations.
MorphNet~\cite{DBLP:journals/corr/abs-1711-06798} can condense the width and depth, but the expanding operation is only able to adapt on width.
Moreover, the expanding operation simply scales the width of all layers uniformly, which highly rely on human prior and not optimal for each layer.
Network Adjustment~\cite{DBLP:conf/cvpr/Chen0XLWT20} can adjust the width of each layer in a non-uniform manner, but it only considers the width dimension and its criterion still rely on human designed rules.
\textit{So can we adapt the network's width and depth automatically in an end-to-end manner?}
In this paper, we proposed a novel network adaption method called \textbf{D}ifferentiable \textbf{N}etwork \textbf{A}daption (DNA). By incorporating Markov modeling with elastic search space, DNA makes the searching process differentiable and automatically adapts the existing network's width and depth to a specified target computation budget.
The gradient-based optimization allows DNA to achieve automatic optimization rather than designs rely on human priors in previous methods.
And the elastic search space allows DNA to achieve the optimization of width and depth in a ``bi-direction" manner.

\begin{figure*}[]
    \centering
    \includegraphics[width=\textwidth]{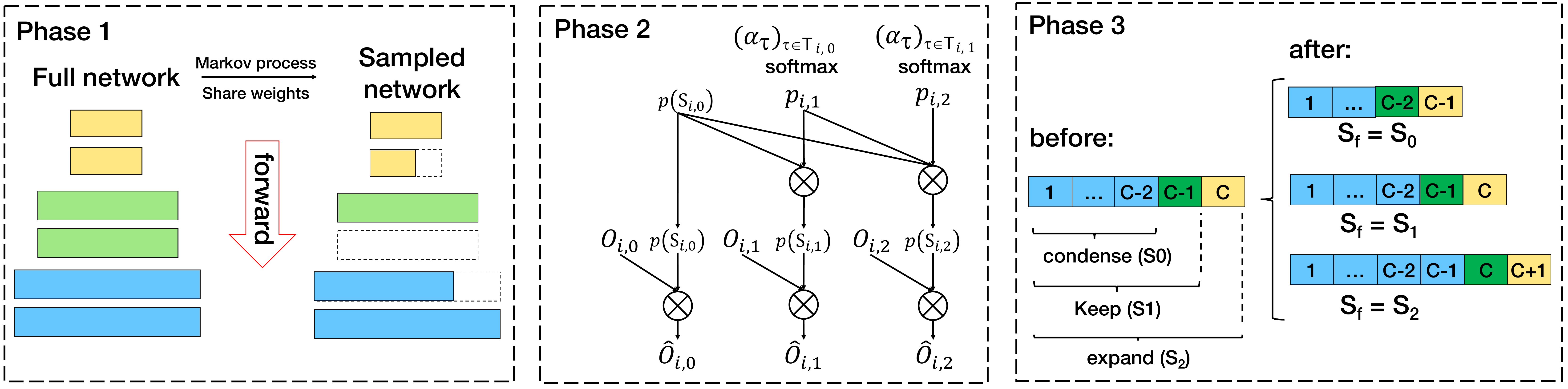}
    \caption{The training pipeline of DNA can be divided into three phases: (1)Train sub-networks sampled from the base network randomly, and update weights of the base network. A wider rectangle indicates that the layer contains more channels, and the hollow parts mean these parts are not included in the sampled network. (2) Fuse the transition probabilities into the base network to optimize the architecture parameters, the notation $p_{i, j}$ is short for $p_\tau = p(S_{i,j}|S_{i,j-1})$. See Section~\ref{sec:diff_markov_modeling} for notations' description. (3) Use the optimized architecture parameters to update the network structure of the base network. The three phases will be called iteratively during training. Best viewed in color.}
    \label{fig:overview}
    \vspace{-1em}
\end{figure*}

In DNA, we first build an elastic search space to support both condensing and expanding operations in the existing network.
Taking the width dimension as an example, we divide the channels in each layer into three parts: condense, keep, and expand, as shown in Figure~\ref{fig:overview} Phase 3. In layer $l$, an adaption procedure is performed to adapt the layer to either of these three parts. 'Expand'  means expanding $l$ by appending extra channels, 'condense' means condense $l$ by removing a certain amount of channels, while 'Keep' means taking no actions. 
The depth of each network stage can also be expanded by appending a new layer or condensed by removing a layer with a similar mechanism.
Then with the elastic search space above, we extend the recently proposed Markov modeling method~\cite{DBLP:conf/cvpr/GuoWLY20} to model the adaption process to support efficient differentiable searching.
We use a set of learnable architecture parameters to represent the transition probabilities of each state, and the adaption procedures can be viewed as transforming from one state to another.
With the proposed method, we successfully achieve network architecture optimization through condensing and expanding in both width and depth dimensions.

Finally, to demonstrate the effectiveness of our method, we conduct exhaustive experiments on ImageNet~\cite{DBLP:journals/ijcv/RussakovskyDSKS15}.
The results proved that DNA can adapt the existing network to meet different requirements of computation budget with better performance than previous works.
What's more, DNA can also improve the performance of existing networks at the same computation budget, even for the already high-accuracy networks obtained by neural architecture search methods such as EfficientNet~\cite{DBLP:journals/corr/HowardZCKWWAA17}.

Our main contributions can be summarized as follows:

\begin{enumerate}
    \item We propose a differentiable searching method for efficient network adaption with human priors as little as possible. As far as we know, this is the first differentiable method for network adaption.
    \item Different from the pre-defined fixed search space in previous neural architecture search methods, we propose a new elastic search space that can flexibly condense or extend, allowing bi-direction optimization for the width and depth of the network. 
    \item Experiments demonstrate the effectiveness of our method, which surpasses previous network adaption works and achieves promising improvements under different computation budgets.
\end{enumerate}


\section{Related Work}

\textbf{Neural architecture search.}
Many works aim to automate the neural architecture design.
Zoph et al.~\cite{DBLP:journals/corr/ZophL16} proposed to use reinforcement learning (RL) to search the structure of neural networks, but these methods need to train a large number of candidate networks from scratch, which requires vast computation resources.
Subsequent works~\cite{DBLP:journals/corr/CaiCZYW17,DBLP:journals/corr/abs-1802-03268} are proposed to improve the efficiency and efficacy of the RL based methods.
DARTS~\cite{DBLP:journals/corr/abs-1806-09055,DBLP:journals/corr/abs-1812-00332} proposed a gradient-based method, which is magnitudes faster than traditional RL based methods.
SPOS~\cite{DBLP:journals/corr/abs-1904-00420} constructs a super-net in which each block has three human-designed choice blocks. It searched a single-path network by evolution algorithm, and the depth and width in it are fixed.
FBNet~\cite{DBLP:journals/corr/abs-1812-03443} incorporates channel searching by searching the expand ratio in each choice block.

The proposed DNA method can be viewed as a gradient-based NAS method that searching the network's depth and width jointly by gradient descent, but there are several major distinctions.
First, the search space of width and depth (i.e. the maximum width and depth) for most NAS methods are mainly designed by human experts and heavily depends on human expert and usually requires large efforts to find the optimal hyper-parameters.
While in DNA, the width and depth can expand or shrink freely, which makes it possible to optimize the width and depth directly instead of finding the optimal solution manually.
Second, gradient-based NAS methods require to construct a super-net that covers the entire search space, thus the super-net is much larger than the candidate networks, which is only applicable to search small networks due to the limitation of memory and computation cost.
While in DNA, the dynamic adaption space makes it not required to construct a super-net that contains all candidate networks, therefore it can be applied to large networks.
Moreover, DNA can be applied to existing networks, either designed by human experts or obtained by NAS methods.

\textbf{Network expansion and network pruning.}
Heuristically methods are direct choices to expand and shrink existing networks.
ResNet~\cite{DBLP:journals/corr/HeZRS15} scales down or up by adjusting the network depth, such as ResNet-50 and ResNet-101. MobileNet~\cite{DBLP:journals/corr/HowardZCKWWAA17,DBLP:journals/corr/abs-1801-04381} uniformly scales the width of each layer to adjust its computation cost.
Most of the previous works only consider one of the operations, expand or shrink, to existing networks.
Net2Net~\cite{DBLP:journals/corr/ChenGS15} proposed a function-preserving transformation to expand the width and depth of existing networks.
AutoGrow~\cite{DBLP:journals/corr/abs-1906-02909} proposed a progressive growing method to expand the depth of the network. 
EfficientNet~\cite{DBLP:journals/corr/abs-1905-11946} uses a compound scaling method to scale up depth, width, and input spatial of a base network by a constant ratio. 
AMC~\cite{DBLP:journals/corr/abs-1802-03494} searches the number of channels in each layer by reinforcement learning.
MetaPruning~\cite{DBLP:journals/corr/abs-1903-10258} searches the number of channels in candidate networks by evolution algorithm.
Network adaption methods are designed for both expanding and shrinking networks at the same time.
MorphNet~\cite{DBLP:journals/corr/abs-1711-06798} prune both channels and layers by sparsifying regularize and expand network width by a heuristic method, i.e. uniformly expand all layers.
Network Adjustment~\cite{DBLP:conf/cvpr/Chen0XLWT20} shrinks or expands each layer's width by pre-defined criterion.

Different from these methods, DNA supports both width and depth shrinking and expanding automatically in a differentiable manner, and it incorporates human priors as little as possible, which reduces the substantial efforts and improves the performance.

\section{Method}
\label{sec:method}

We propose a network adaption method called DNA, which can adapt a given network to specified computation budget, while does not require to maintain a large super-network like NAS methods. The pseudo-code of the entire method is given in Algorithm~\ref{algo:dna}.
First, the network is partitioned into an elastic search space, in which each layer/stage can only be manipulated by one unit, where ``unit'' means certain amount of channels or layers, and the details are given in Section~\ref{sec:adpation_space}, then in each adaption step $i$, the intermediate network $N_i$ is extended as a super-network and trained by sampling sub-networks from elastic space, which will be introduced in Section~\ref{sec:training_subnet}, after training $N_i$, we model the adaption process by a set of learnable architecture parameter $A_i$, which can be fused into network $N_i$ and optimized end-to-end by gradient descent, and 
then, we use optimized $A_i$ to perform the network adaption by Markov process, we will explain the whole procedure in Section~\ref{sec:arch_param}.





\begin{algorithm}
\SetAlgoLined
\SetKwInOut{Input}{input}
\SetKwInOut{Output}{output}

\Indm  
  \Input{network $\mathbf{N}$, budget $\mathbf{C}$, dataset $\mathbf{D}$, Arch params $\mathbf{A}_0$}
  \Output{optimized network $\mathbf{N}_T$ that meeting the budget}
\Indp
 
\For{i = 1, 2, ... T} {
    $\mathbf{N_{i-1}}$ = PartitionElesticSearchSpace($\mathbf{N_{i-1}}$)
    
    \For{$b$ in $\mathbf{D}$} {
     \tcc{$b$ indicates one batch}
         \For{$k$ = 1, 2, ..., M} {
            $\mathbf{n}_k$ = SampleSubNetworks($\mathbf{N_{i-1}}$)
        
            TrainNet($\mathbf{n}_k$, $b$)
        }
        TrainNet($\mathbf{N_{i-1}}$, $b$)
        
        $\mathbf{n_{min}}$ = SampleMinNetworks($\mathbf{N_{i-1}}$)
        
        TrainNet($\mathbf{n_{min}}$, $b$)
    }
    $\mathbf{A_i}$ = TrainArchParams($\mathbf{N_{i-1}}$, $\mathbf{A_{i-1}}$, $\mathbf{C}$, $\mathbf{D}$)
    
    $\mathbf{N_i}$ = UpdateNetStruct($\mathbf{N_{i-1}}$, $\mathbf{A_i}$)
}

TrainNet($\mathbf{N}_T$, $\mathbf{D}$)

\textbf{return} $N_T$
 \caption{Differentiable Network Adaption}
 \label{algo:dna}
\end{algorithm}

\subsection{Elastic Search Space}
\label{sec:adpation_space}

We perform network adaption by manipulate the network width and depth. We first explain the single dimension space respectively and then combine together to form a joint space. 

\textbf{Width space.} 
In most widely used convolutional neural networks, a convolutional layer could have thousands of channels, which is extremely hard to perform a fine-grained searching.
To further reduce the complexity of search space, we divide the channels into $K$ channel groups such that each group has an equal amount of channels.
and our adaption method allows each layer to be expanded or condensed by only one group in each step.

In layer $l$, the channel groups is given by $G = \{g_1, g_2, ... g_K\}$, therefore the condense operation can be represented as :$G\prime = G \setminus \{g_K\}$, and the expansion operation is performed by adding an extra group $g_{K+1}$ to $G$ such that $G\prime = G \cup \{g_{K+1}\}$, where $G\prime$ denote the updated channel groups of layer $l$.

\textbf{Depth space.} 
The network adaption in depth dimension is performed by manipulating the number of layers in each stage of CNN. In a network stage $D$ that has $L$ layers, similarly as width space, we denote $D \setminus \{d_{L}\}$ and $D \cup \{d_{L+1}\}$ as the condense and expansion operations respectively.


\textbf{Joint space.}
In a stage $D$, the last layer is represented by $d_L$, which has only one channel group left, denoted as $G_L = {g_1}$, therefore, $D \setminus \{d_L\}$ in depth dimension and $G_L \setminus \{g_1\}$ in width dimension will be the same operation. In order to unify the adaption operations in joint space, we introduce a constraint that in width dimension, each layer must retain at least one group and it can only be condensed by depth operation. 

\textbf{Greedy adaption.}
Denoted the initial network as $N_0$,
$N_0 \rightarrow N_1 \rightarrow ... \rightarrow N_T$ represents our adaption process that greedily adapt $N_0$ to target network $N_T$.
while in each step, the $N_{i+1}$ is assume to be the optimal network in the elastic search space of $N_i$. The detail of adaption process is introduced in Section~\ref{sec:network_adaption_operation}.

\begin{figure}
    \centering
    \includegraphics[width=0.45\textwidth]{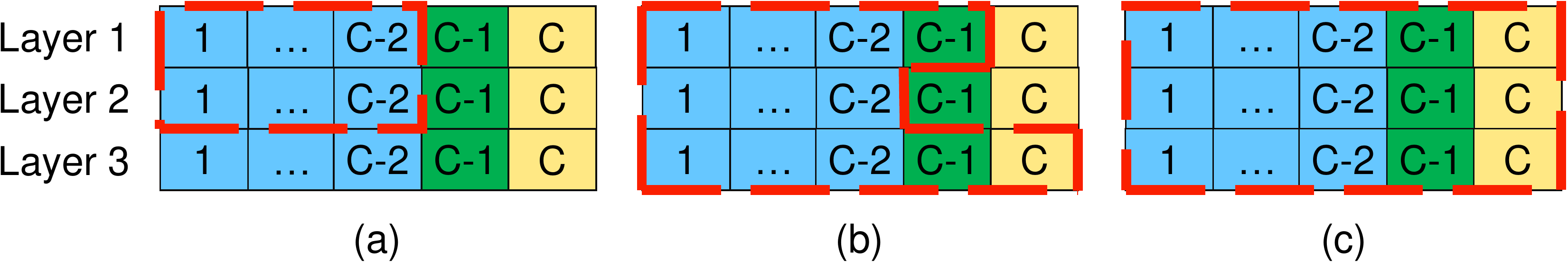}
    \caption{A three-layer network sample for illustrating the base network sampling in Phase 1 of training. Each layer has C channels. The red dashed line surrounding parts represent the sampled network. (a) is the minimum possible network. (b) is a randomly sampled network. (c) is the base network or super-net. All the sampled networks share their weights.
    \vspace{-1em}}
    \label{fig:subnet_sample}
\end{figure}

\subsection{Training of super-network.}
\label{sec:training_subnet}
In each step $i$, the network $N_i$ can be viewed as a super-net that contains sub-networks and all sub-networks share common weights. 
sub-networks are sampled by randomly in joint space, and the gradient of all sub-networks are accumulated to the weights of the super-net.
This process aims to approximate the performance of sub-networks trained with standalone weights so that their performances are not affected much by the varying super-net architectures. However, one question may raised: \textit{how do we know the performance improvement in advance if the network is expanded?}
We tackle this problem by extend each layer by one group and each stage by one layer respectively, by such formulation, each layer $L$ of sub-networks has three possible choice: $\{K_L-1, K_L, K_L+1\}$ and the same in each stage $D$: $\{L_D-1, L_D, L_D+1\}$, therefore, the sample space has the complexity of $3^{(n_l+n_s)}$, where $n_l$ and $n_s$ denote the number of layer and the number of stage in $N_i$ respectively. 
We extend the ``sandwich rule'' proposed in ~\cite{DBLP:journals/corr/abs-1903-05134} that besides training the sub-networks, we also train the full base network to guarantee sufficient training of all network weights, and train the smallest network in joint space to pull up the minimum accuracy among entire network, as shown in Figure~\ref{fig:subnet_sample}. 
So the loss function $loss_{task}$ in Phase 1 is the sum of three cross-entropy loss: the loss for sampled sub-networks $loss_{sampling}$ and the loss for full network $loss_{full}$, and the loss for minimum network $loss_{min}$
\begin{align}
\label{eq:task_loss}
loss_{task} = loss_{sampling} + loss_{full} + loss_{min}
\end{align}

\subsection{Modeling architecture parameters.}
\label{sec:arch_param}
\begin{figure}[t!]
    \centering
    \includegraphics[width=0.45\textwidth]{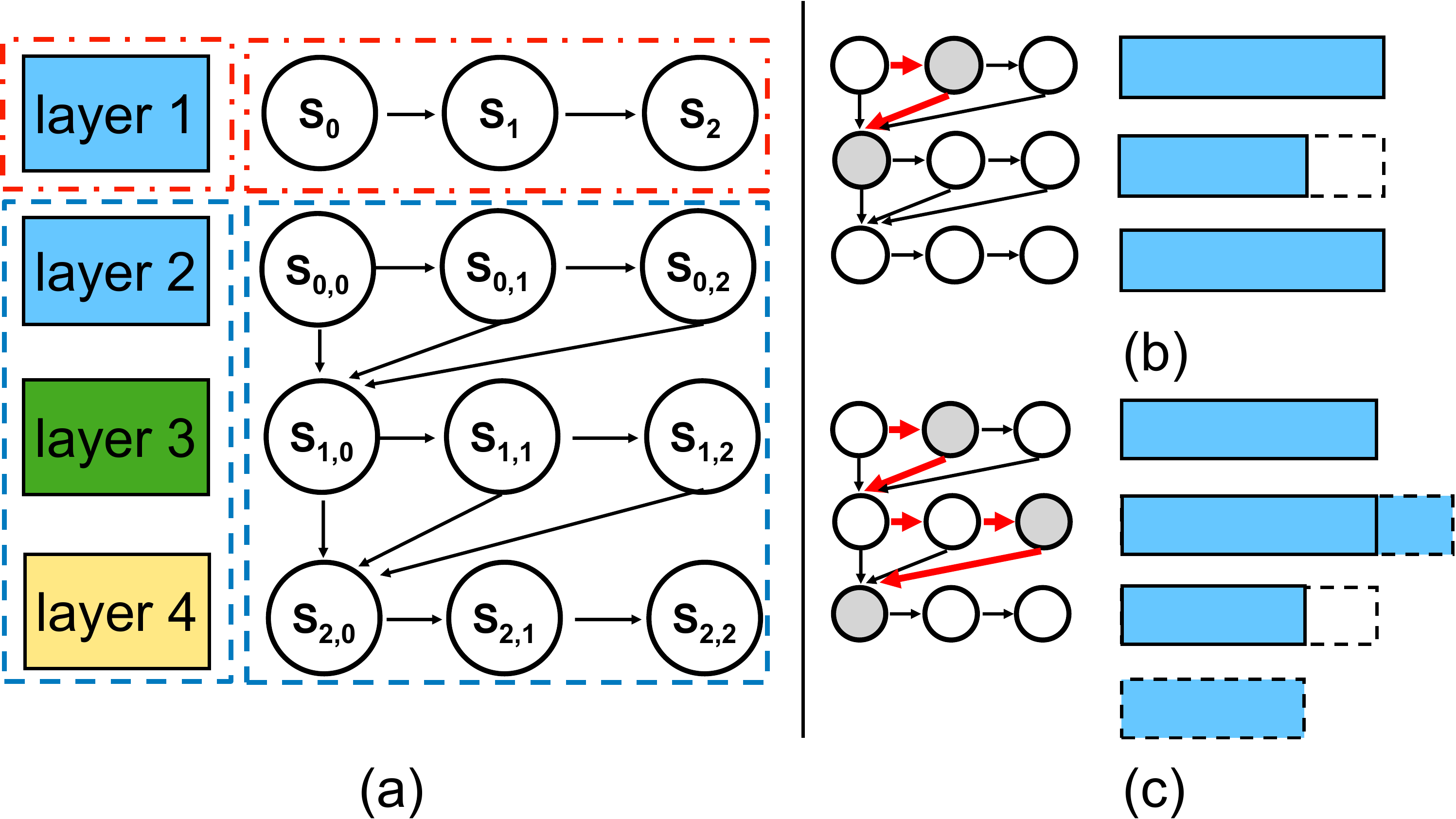}
    \caption{(a) left parts show a depth space. The right parts show the corresponding joint space formulation. The layer within $S_0$ (e.g. layer 1) only has width space, while the rest of all layers formulate a joint space. The arrows indicate the valid transitions from one state to another. (b) and (c) give two examples of how the adaption process is modeled by a sequence of transitions. Red arrows indicate the transition procedure, and the last state of a layer, i.e. the solid gray circle, will determine the operation on itself.
    \vspace{-1em}}
    \label{fig:markov}
\end{figure}

To allow the adaption process be optimized end-to-end by gradient descent, we extend the Markov modeling method proposed in~\cite{DBLP:conf/cvpr/GuoWLY20} to model the adaption operations as state transitions, which is shown Figure~\ref{fig:markov}, in which the transitions are modeled by a set of learnable architecture parameters. 

The transitions over joint space are shown in the left parts of Figure~\ref{fig:markov}(a).
The horizontal transitions determine the width space in each layer, and the vertical transitions determine the depth space in each stage. 
In horizontal transitions, $P(S_{i,j+1}|S_{i,j})$ denotes the transition probability in width space from $S_{j}^{width}$ to $S_{j+1}^{width}$ within depth state $S_i^{depth}$, while in vertical transitions,
$P(S_{i+1, 0}|S_{i,j})$ is the transition probability from $S_j^{width}$ in  $S_{i}^{depth}$ to $S_{i+1}^{depth}$. Therefore, the initial state $S_{0,0}$ denote the minimum network and $S_{2,2}$ denote the full network, a sequence of transitions starts from $S_{0,0}$ determines the number of layer and the number of channel group each layer simultaneously, i.e. determines a sub-network in the elastic space.
Besides, each state has an extra transition to terminal state $S_E$ to indicate that the adaption process is ended, and the transition probability is denoted as $p(S_E|S_{i,j})$.


\subsubsection{Optimize architecture parameters.}
\label{sec:diff_markov_modeling}
In each layer, we denote the channel groups as $\{g_1, g_2, ..., g_{K+1}\}$ in which $g_i$ is $i^{th}$ channel groups. We define the marginal probability that first $K-1$ groups are included in sub-network as $p(g_1, ..., g_{k-1})=p(S_0)$, similarly, we use $p(S_i)$ to represent the marginal probability of the channel groups in $S_i$ are included in the sub-network, which can be computed by:
\begin{equation}
\begin{aligned}
\label{eq:marginal_prob}
    p(S_{j}) = p(S_{0})\prod_{k=1}^{j}p(S_{k}|S_{k-1})
\end{aligned}
\end{equation}
where the $p(S_0) = 1$ in width space. While in joint space, $p(S_{i, 0})(i>0)$ is the marginal probability of $p(S_0^{width})$ in depth state $i$, which can be computed as follows:
{\small
\begin{align}
\label{eq:e_channel}
p(S_{i, 0}) = \sum_{j=0}^{2}p(S_{i, 0}|S_{i-1, j})p(S_{i-1,j})
\end{align}
}
Let $p_\tau = \frac{exp^{(-\alpha_\tau)}}{\sum_{\tau' \in T_j}exp^{(-\alpha_\tau')}}$ to represent the transitions probability $p(S_i|S_j)$, and $T_{j}$ is the set of all transitions from $S_j$ any other states (including terminal state $S_E$), $\alpha_\tau$ is the corresponded learnable architecture parameters of transition $\tau \in T_j$.

In layer $i$, the channels (i.e. convolutional filters) are divided into 3 states, given input $x^{(i)}$, the output $O_{i,j}$ of each state $S_{i,j}$ is computed by:
\begin{align}
\label{eq:conv}
O_{i,j} = w_{i,j} \odot x^{(l)}
\end{align}
where $w_{i,j}$ is the network weights in $S_{i,j}$ and $\odot$ denote the convolution operations.
Then the learnable parameters $\alpha$ are wrapped into the base network by equation:
$\hat{O_{i,j}} = O_{i,j} \times p(S_{i, j})$, in which $\hat{O_{i,j}}$ is the actual output of state $S_{i,j}$. Therefore the architecture parameters can be optimized end-to-end by gradient descent.

\subsubsection{Computation Budget Regularization}
\label{sec:regularization}
In this section, we will introduce how to regularize the computation budget.
First, we will explain how to compute the expected network width by transition probabilities mentioned in Section~\ref{sec:diff_markov_modeling}. With the expected width, the expected computation cost of the network can be computed.

For a layer $i$, the expected channel $E(channel)$ is computed by:
\begin{align}
\label{eq:e_channel}
E(channel) = \sum_{j=0}^{2}p(S_{i,j})N_{c}(S_{i,j}) 
\end{align}
where $N_{c}(S_{i,j})$ is the number of channels in state $S_{i,j}$. 
Note that the probability that layer $i$ exists in the networks is $p(S_{i, 0})$, which is computed Equation~\ref{eq:e_channel}. Therefore, we do not need to compute expected number layers separately.


\label{sec:budget_reg}
In this work, we use FLOPs (floating-point operations) to measure the computation cost of the model, and FLOPs can be replaced by other measurements such as inference latency. 
In layer $l$, the expected output channels $E(out)$ can be computed by Equation~\ref{eq:e_channel}, while the expected input channel $E(in)$ is the output channel of $(l-1)^{th}$ layer, and the expected FLOPs $E(F^{(l)})$ can be computed by:
\begin{eqnarray}
\label{eq:e_fl_layer}
E(F^{(l)}) =& E(out)E(in)(F_{Conv})
\end{eqnarray}
where $F_{Conv}$ is the FLOPs of one convolution operation over the feature map.
Then the expected FLOPs of a $N$-layer network $E(N_{F})$ is:
\begin{align}
\label{eq:e_flops}
E(N_{F}) = \sum_{l=1}^{N}E(F^{(l)})
\end{align}
in which N is the number of convolutional layers.
Given target computation budget $F_T$, we formulated the computation budget regularization loss $loss_{F}$ as follows:
\[
\displaystyle loss_{F} = 
\begin{cases}
\label{eq:flops_loss}
 0 & \text{,$E(N_{F}) \leq F_T$} \\ 
log(|E(N_{F}) - F_T|) & \text{,$E(N_{F}) > F_T$} \\ 
\end{cases}
\]
With the above equation, we can optimize the expected FLOPs of a network by gradient descent.
And the loss function $loss_M$ of optimizing the Markov model is:
\begin{align}
\label{eq:upd_a_loss}
loss_M = loss_{task} + \lambda_{F} loss_{F}
\end{align}
$\lambda_F$ is a hyper-parameter adjusted by users. 
Note that only architecture parameters are updated in Phase 2.

\subsection{Network Adaption Operation}
\label{sec:network_adaption_operation}
At the end of step $t$, the network will perform an adaption operation, i.e. adapt the width of each layer and the depth of each stage.
Instead of sampling in the Markov model, we start from $S(0)$ or $S(0,0)$ and directly select the transition with maximum probability, as shown in Figure~\ref{fig:markov} (b)(c).
The last state $S_f$ of each layer, solid gray circle, will determine the operation on itself.
For layers that only update in the width space:
(1) If $S_f = S_0$, it means the layer tends to retain fewer channels, therefore, the $(K_t)^{th}$ group will be removed and $K_{t+1} = K_t-1$.
(2) If $S_f = S_1$, we can infer this layer tends to maintain the current width, thus no groups will be added or removed.
(3) If $S_f = S_2$, it indicates that this layer tends to expand more channels. Therefore, a new group will be appended at the end of the layer, resulting in $(K_{t}+1)$ groups.
For layers that update in the joint space, the $S_f$ will be decomposed into a single dimension and perform a similar operation as above.

Note that the order of channels does not change the structure of the network, channels are removed or added at the end of each layer for the purpose of reducing the complexity of search space.
Similar situation with it in depth dimension, we only remove or append layers at the end of each stage as the order of layers does not change the structure of the network either.
For the newly added channel group or layer, the weights are randomly initialized. And the size of the newly added layer is the same as its previous layer.

\subsection{Obtaining target network.}
\label{sec:obtain_final}
After DNA's training is done, we compute a ``expected network'' as the final network.
In each stage of expected network, the number of layers $E(layer)$ is computed by:
\begin{align}
\label{eq:e_depth} 
E(layer) = N_l(S_{0,*}) + \sum_{j=1}^{2}p(S_{j, 0}) 
\end{align}
where $N_l(S_{0,*})$ denote the number of layers that assigned to depth state $S_0$, and
$p(S_{j, 0})$ is the marginal probability of $S_{j, 0}$ computed in Equation~\ref{eq:marginal_prob}, and we use $round(E(layer))$ as the number of retained layers in corresponding stage. In each retained layer, $round(E(channel))$ is the number of channels of this layer and $E(channels)$ is computed by Equation~\ref{eq:e_channel}.
The experiments show that the computation cost of the final ``expected model'' can always converge to the target computation budget, and the performance is more stable compared with networks by Markov process over the joint search space.
Then the final target network is trained from scratch.

\section{Experiments}
In this section, we conduct extensive experiments on different architectures to verify the effectiveness of DNA.

\begin{figure}
    \includegraphics[width=0.45\textwidth]{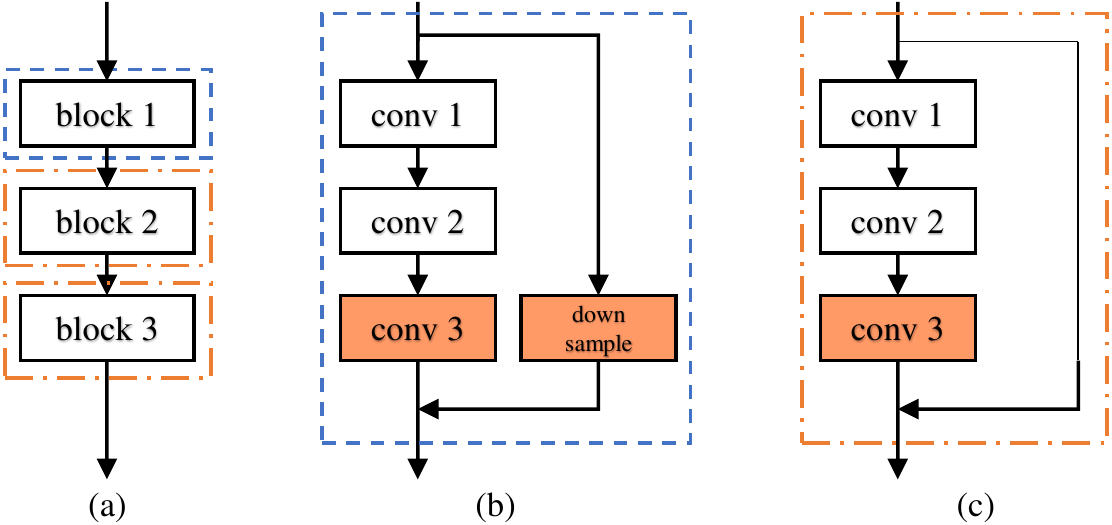}
    \caption{Shortcut case. (a) is a stage with three blocks. (b) is the structure of block 1. (c) is the structure of block 2 and block 3. Colored parts in (b) and (c) means they share the same state.}
    \label{fig:residual_case}
\end{figure}

\begin{table*}[h!]
\centering
\caption{The performance of network adaption on different dimensions in $F_T \approx F_S$ scenario, adjusting base network under the same computation budget. ${\uparrow}$ means relative improvement compared with baseline.}
\begin{tabular}{ccc|cc|cc|cc|cc}\\ 
\toprule
 &                &                & \multicolumn{8}{c}{base network $\veryshortarrow$ target network}\\
 \cmidrule{4-11}
 & \multicolumn{2}{c|}{dimension} &\multicolumn{2}{c|}{MobileNet-v2}&\multicolumn{2}{c|}{EfficientNet-B0} & \multicolumn{2}{c|}{ResNet-18} & \multicolumn{2}{c}{ResNet-50} \\
 \cmidrule{2-11}
                        & width & depth & FLOPs & Top-1 & FLOPs & Top-1 & FLOPs & Top-1 & FLOPs & Top-1 \\
\midrule
Baseline                &  &            & 300M & 72.8 & 385M & 76.2 & 1.8G & 70.3 & 4.1G & 76.7 \\
\midrule
\multirow{3}{*}{DNA}   
                        & \checkmark &            &  300M & $73.4_{\uparrow0.6}$ & 385M & $76.3_{\uparrow0.1}$ & 1.8G & $71.3_{\uparrow1.0}$ & 4.1G & $77.2_{\uparrow0.5}$ \\
                        &  & \checkmark           &  300M & $73.1_{\uparrow0.3}$ & 385M & $76.4_{\uparrow0.2}$ & 1.8G & $70.8_{\uparrow0.5}$ & 4.1G & $77.0_{\uparrow0.3}$ \\
                        & \checkmark & \checkmark &  295M & $73.6_{\uparrow0.8}$ & 385M & $76.7_{\uparrow0.5}$ & 1.8G & $71.7_{\uparrow1.4}$ & 4.1G & $77.6_{\uparrow0.9}$ \\
\bottomrule
\end{tabular}

\label{tb:transform}

\end{table*}

\begin{table*}[h!]
\centering
\caption{The performance of network adaption on different dimensions in $F_T > F_S$ scenario, adjusting the base network to a larger computation budget. * means EfficientNet-B0 with the same input resolution as EfficientNet-B1. ${\uparrow}$ means relative improvement compared with baseline.}
\begin{tabular}{ccc|cc|cc|cc|cc}\\ 
\toprule
 &                &                & \multicolumn{8}{c}{base network $\veryshortarrow$ target network}\\
 \cmidrule{4-11}
 & \multicolumn{2}{c|}{dimension} &\multicolumn{2}{c|}{MBV2 1.0$\times$ $\veryshortarrow$ 1.4$\times$}&\multicolumn{2}{c|}{Efficient-B0* $\veryshortarrow$ B1} & \multicolumn{2}{c|}{R18 $\veryshortarrow$ R34} & \multicolumn{2}{c}{ResNet-50} \\
 \cmidrule{2-11}
                        & width & depth & FLOPs & Top-1 & FLOPs & Top-1 & FLOPs & Top-1 & FLOPs & Top-1 \\
\midrule
Baseline                &  &            & 580M & 75.7 & 685M & 77.4 & 3.7G & 74.0 & 7.8G & 78.2 \\
\midrule
\multirow{3}{*}{DNA}   
                        & \checkmark &            &  580M & $76.1_{\uparrow0.4}$ & 685M & $77.9_{\uparrow0.5}$ & 3.7G & $74.3_{\uparrow0.3}$ & 7.8G & $78.6_{\uparrow0.4}$ \\
                        &  & \checkmark           &  580M & $75.9_{\uparrow0.2}$ & 680M & $78.0_{\uparrow0.6}$ & 3.7G & $74.4_{\uparrow0.4}$ & 7.8G & $78.5_{\uparrow0.3}$ \\
                        & \checkmark & \checkmark &  580M & $76.2_{\uparrow0.5}$ & 679M & $78.2_{\uparrow0.8}$ & 3.7G & $74.6_{\uparrow0.6}$ & 7.8G & $78.8_{\uparrow0.6}$ \\
\bottomrule
\end{tabular}
\label{tb:expansion}
\end{table*}

\begin{table*}[h]
\centering
\caption{The performance of network adaption on different dimensions in $F_T < F_S$ scenario, adjusting the base network to a smaller computation budget. ${\uparrow}$ means relative improvement compared with baseline.}
\begin{tabular}{ccc|cc|cc|cc|cc}\\
\toprule
 &                &                & \multicolumn{8}{c}{base network $\veryshortarrow$ target network}\\
 \cmidrule{4-11}
 & \multicolumn{2}{c|}{dimension} &\multicolumn{2}{c|}{MBV2 1.0$\times$ $\veryshortarrow$ 0.35$\times$ } & \multicolumn{2}{c|}{MBV2 1.0$\times$ $\veryshortarrow$ 0.75$\times$}& \multicolumn{2}{c|}{R18 1.0$\times$ $\veryshortarrow$ 0.85$\times$} & \multicolumn{2}{c}{R50 1.0$\times$ $\veryshortarrow$ 0.85$\times$} \\
 \cmidrule{2-11}
                        & width & depth & FLOPs & Top-1 & FLOPs & Top-1 & FLOPs & Top-1 & FLOPs & Top-1 \\
\hline
Baseline                & &            & 59M & 60.3 & 210M & 70.4 & 1.08G & 67.5 & 3.0G & 75.3 \\
\hline
\multirow{3}{*}{DNA}   
                        & \checkmark &            &  59M & $62.9_{\uparrow2.6}$ & 210M & $71.6_{\uparrow1.2}$ & 1.08G & $68.2_{\uparrow0.7}$ & 3.0G & $76.2_{\uparrow0.9}$ \\
                        &  & \checkmark           &  59M & $61.5_{\uparrow1.2}$ & 210M & $70.8_{\uparrow0.4}$ & 1.08G & $68.0_{\uparrow0.5}$ & 3.0G & $75.9_{\uparrow0.6}$ \\
                        & \checkmark & \checkmark &  59M & $62.4_{\uparrow2.1}$ & 210M & $72.4_{\uparrow2.0}$ & 1.08G & $68.5_{\uparrow1.0}$ & 2.8G & $76.9_{\uparrow1.6}$ \\
\bottomrule
\end{tabular}

\label{tb:pruning}
\end{table*}

\subsection{Implementation Details}
\label{sec:setting}
We show the effectiveness of DNA on ImageNet classification~\cite{DBLP:journals/ijcv/RussakovskyDSKS15} which contains 1000 classes. We perform experiments on both human-designed networks (MobileNet-v2~\cite{DBLP:journals/corr/abs-1801-04381}, ResNet~\cite{DBLP:journals/corr/HeZRS15}) and the networks searched by the NAS method (EffcientNet~\cite{DBLP:journals/corr/abs-1905-11946}, MobileNet-v3~\cite{DBLP:journals/corr/abs-1905-02244}).

In width dimension, we equally divide channels in each layer into 10 groups.
All base network structures used in DNA are the same as in their original paper. In the following experiments, we will explain the detailed setting of each base network.

\textbf{Training of DNA.}
The training is performed on 16 Nvidia GTX 1080TI GPUs with a batch size of 1024.
As explained in Section~\ref{sec:method}, DNA's training pipeline contains three phases, and three phases are called iteratively during the training process. In all experiments, we train phase 1 for 1.5 epochs and train phase 2 for 0.5 epoch. Three phases are executed 25 times, resulting in 50 epochs in total. The network's weights are trained by stochastic gradient descent (SGD). The initial learning rate is 0.2 and is reduced to 0.02 by cosine annealing. The architecture parameters are trained by the ADAM optimizer~\cite{kingma2014adam:} with a learning rate of $0.01$. 
The loss weight $\lambda_{F}$ is set to 0.1 in all experiments.

\textbf{Warmup pre-training of base network.} 
To prevent the adaption from trapping into the local minima at the beginning of the training, we add a warmup pre-training before training DNA. In the warm-up pre-training, the base network is trained by only running phase 1 for 15 epochs.

\textbf{Training of the target network.}
The target networks are trained from scratch on 16 GPUs with a batch size of 1024. In our experiments, ResNet is trained for 100 epochs, MobileNet-v2 and EfficientNet are trained for 200 epochs. All networks are trained by stochastic gradient descent with an initial learning rate of 0.4 and decay to 0 by cosine annealing. 
When training EfficientNet, we modify the settings in the original paper~\cite{tan2019efficientnet:} to adopt drop connect with ratio 0.2. No other training enhancements are used (e.g. AutoAugment~\cite{DBLP:journals/corr/abs-1805-09501} and EMA~\cite{tan2019efficientnet:}).

\textbf{Base network with shortcut}
Most of the widely used networks are constructed by a stack of blocks and in each stage, all block has several layers that share the same topology. Thus removing or adding a single layer inside the block is not feasible. For example, in MobileNet-v2, we cannot remove the $1\times1$ point-wise convolutional layer in each linear bottleneck block as it will change the topology of this block. Therefore in the depth 
dimension of these networks, we adjust the number of blocks instead of layers.

Some blocks have identity shortcut connections (e.g. ResNet, MobileNet-v2).
The input and output width in such a block must be the same as the output width of its previous block. 
Figure~\ref{fig:residual_case} shows a case of a stage with 3 blocks in ResNet, in which the second and third blocks have identity shortcut, and the first block has a down-sample shortcut. 
In this case, the input width of conv1 and output width of conv3 in the last two blocks will share the same state as the output width of conv3 in the first block.

\subsection{Ablation Studies}
\label{sec:ablation}

\subsubsection{Network Adaption on different dimensions}
In this section, we perform experiments to verify the influence of different adaption dimensions in different cases.

The \textbf{base networks} used in following experiments are MobileNet-v2 1.0$\times$, EffcieintNet-B0, ResNet-18 and ResNet-50. And $\alpha\times$ in MobileNet-v2 means the number of channel in each layer is uniformly scaled by $\alpha$.
All \textbf{baselines}, i.e existing networks with target computation budget, are trained with the same setting in Section~\ref{sec:setting}.
Since DNA is a flexible method to adapt networks with computation cost $F_S$ to different target computation budget $F_T$, to demonstrate the effectiveness and generalization ability of DNA, we conduct experiments on three major scenarios which are described as follows: 
\begin{itemize}
    \item ($F_T \approx F_S$). Optimize networks under the same computation budgets. The results are reported in Table~\ref{tb:transform}. Noted that the baselines are the base networks, i.e. MobileNe-v2, EfficientNet-B0, ResNet-18 and ResNet-50. 
    \item ($F_T > F_S$):  Adapt the base network to a larger computation budget, i.e. expend the model. The results are reported in Table~\ref{tb:expansion}. The baseline models are MobileNet-v2 1.4$\times$, EfficientNet-B1, ResNet-34 and ResNet-101, which are the design from their original papers~\cite{DBLP:journals/corr/abs-1801-04381, DBLP:journals/corr/HeZRS15, DBLP:journals/corr/abs-1905-11946}. 
    \item ($F_T < F_S$):  Adapt the base network to a smaller computation budget, i.e. condense the model. The results are reported in Table~\ref{tb:pruning}. The baseline models are MobileNet-v2 0.75$\times$, MobileNet-v2 0.35$\times$ and ResNet-50 0.85$\times$. 
\end{itemize}
Among the three cases, searching on both width and depth adaption space can improve the performance, and the performance can further be improved by the compound space.
What's more, the performance can be further improved even for a state-of-the-art searched architecture EfficientNet, which demonstrates the effectiveness of DNA on network adaption.





\subsubsection{Influence of different network obtaining methods}
In Phase 3 of the DNA pipeline, we obtain the target network by expectation. To evaluate the influence of different final network obtaining: expected network and network by Markov sampling, we sample 5 target networks within target computation budgets and train them from scratch.
We use ResNet-50 in the ($F_S \approx F_T$) scenario.
The results are listed in Table~\ref{tb:effective_markov}.
The results show that the target network obtained by the expected network achieved the highest performance, and the performance of networks sampled by the Markov process slightly worse than it.

\begin{table}[H]

\centering
\caption{Comparison of different final network obtaining methods.}
\begin{tabular}{c|c|cc}
\toprule
\multirow{2}{*}{Model} & \multirow{2}{*}{Expected} &\multicolumn{2}{c}{Markov} \\ 
                      &                  &  Max  & Min     \\
\midrule
ResNet-50 4.1G & 77.6   & 77.4  & 76.7    \\ 
\midrule
\end{tabular}

\label{tb:effective_markov}
\end{table}

\subsubsection{Recoverability of the searching method}
\begin{figure}[t]
  \centering
  \includegraphics[width=0.4\textwidth]{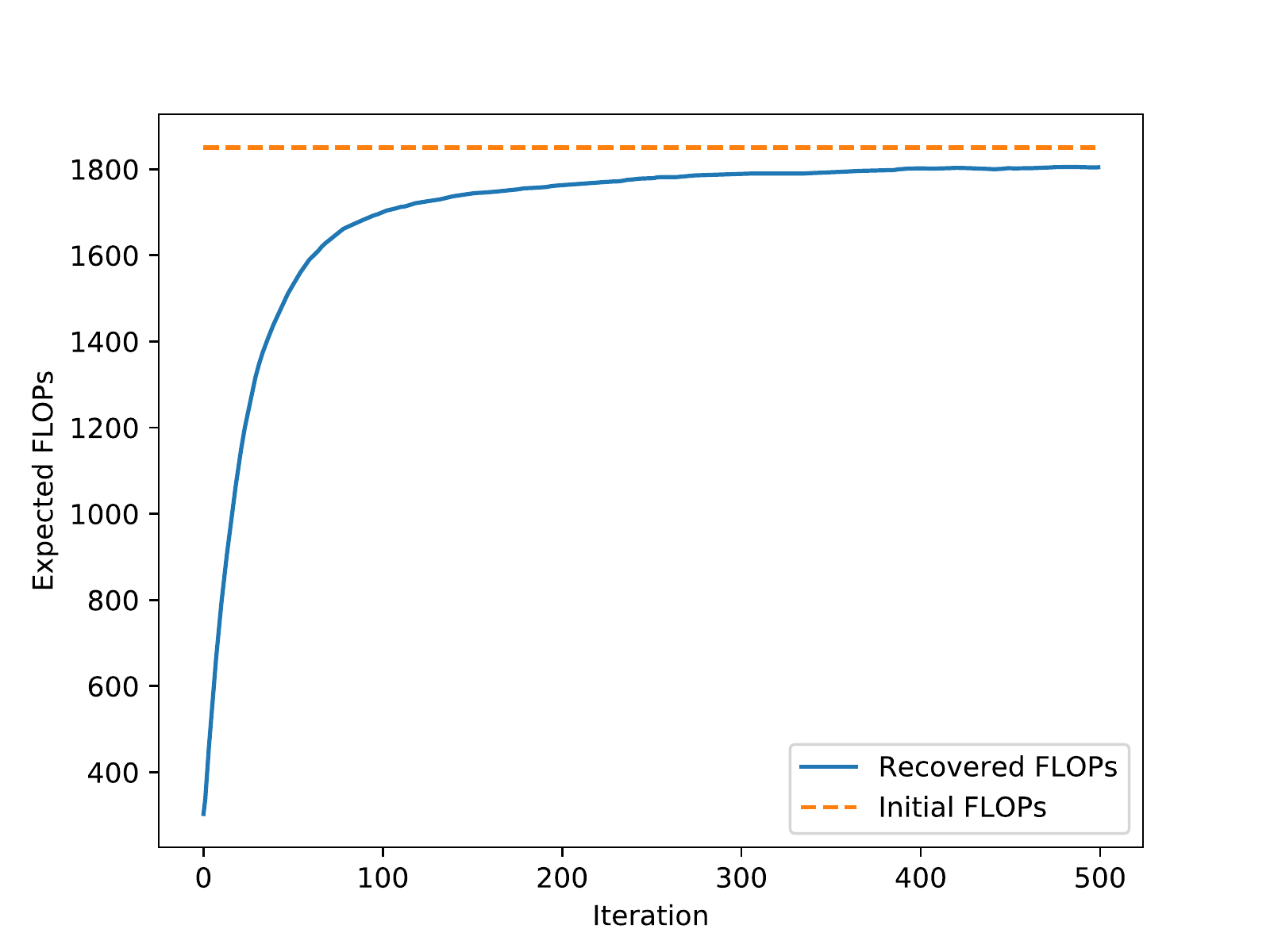}
  \caption{Recoverability of the searching method in DNA.}
  \label{fig:recover}
  \vspace{-1.5em}
\end{figure}

In this section, we verify the recoverability of the searching method in DNA, 
which refers to the property  our method should have to retain nearly all groups and layers when adapting the base network without computation budget regularization. We use a pre-trained ResNet-50 as the base network with randomly initialized learnable parameters in the Markov model. We freeze the weights of the ResNet-50 and only running phase 2 in the DNA pipeline to optimize the Markov model only with the task loss. The result in Figure~\ref{fig:recover} shows that the FLOPs of our method can recover to those of the pre-trained model within certain iterations.

\subsubsection{Influence of different initial base networks}
In this section, we evaluate the influence of different base networks for network adaption to the same target computation budget.
In the experiments, we adapt different ResNet~\cite{DBLP:journals/corr/HeZRS15} models to same target FLOPs.
Note that the block structure in all base models is the same.
The results in Table~\ref{tb:robust} show that given the same computation budget, our method is able to adapt different models (with same block structure) into target FLOPs with the same performance, which shows the robustness of DNA.
\begin{table}[H]
\centering
\caption{Same target FLOPs with different base model.}
\begin{tabular}{lc|cc}
\toprule
base model & initial FLOPs &  target FLOPs & Top-1\\
\midrule
Res18 0.85x & 1.08G & 1.8G & 71.6 \\
Res18 1.0x & 1.8G & 1.8G & 71.6 \\
Res34 1.0x & 3.7G & 1.7G & 71.4\\

\bottomrule
\end{tabular}

\label{tb:robust}
\end{table}

\begin{table}[t]
\centering
\caption{Comparison of state-of-the-art methods under different FLOPs settings in ResNet.}
\begin{tabular}{l|l|lll}
\toprule
base net                                       & FLOPs                  & method        & Top-1 & $\Delta$ \\ \midrule

\multicolumn{1}{c|}{\multirow{7}{*}{ResNet18}} & \multirow{1}{*}{1.8G}  & baseline       & 70.3  & -     \\ \cmidrule{2-5}
\multicolumn{1}{c|}{}                          & \multirow{2}{*}{1.08G} & Uniform 0.85$\times$       & 67.5  & -2.8     \\
\multicolumn{1}{c|}{}                          &                        & \textbf{DNA}           & \textbf{68.5}  & \textbf{-1.8}     \\ \cmidrule{2-5} 

\multicolumn{1}{c|}{}                          & \multirow{2}{*}{1.8G}  & Adjustment~\cite{DBLP:conf/cvpr/Chen0XLWT20}           & 71.2  & +0.9     \\
\multicolumn{1}{c|}{}                          &                        & \textbf{DNA}           & \textbf{71.6}  & \textbf{+1.3}     \\ \cmidrule{2-5} 

\multicolumn{1}{c|}{}                          & \multirow{2}{*}{3.7G}  & R18 $\veryshortarrow$ R34 & 74.0  & +3.7        \\
\multicolumn{1}{c|}{}                          &                        & \textbf{DNA}           & \textbf{74.6}  & \textbf{+4.3}     \\ \midrule
                         
\multirow{14}{*}{ResNet50}                     & \multirow{1}{*}{4.1G}  & baseline       & 76.6  & -     \\ \cline{2-5}
                                               & \multirow{5}{*}{2.3G}  & Uniform 0.75$\times$      & 74.6  & -2.0     \\
                                               &                        & Meta~\cite{DBLP:journals/corr/abs-1903-10258}          & 75.4  & -1.2     \\
                                               &                        & JMP~\cite{DBLP:journals/corr/abs-2005-08931}           & 75.6  & -1.0     \\
                                               &                        & \textbf{DMCP}~\cite{DBLP:conf/cvpr/GuoWLY20}          & \textbf{76.2}  & \textbf{-0.4}     \\
                                               &                        & \textbf{DNA}           & \textbf{76.2}  & \textbf{-0.4}     \\ \cmidrule{2-5} 

                                               & \multirow{5}{*}{3.0G}  & Uniform 0.85$\times$      & 75.4  & -1.2     \\
                                               &                        & Meta~\cite{DBLP:journals/corr/abs-1903-10258}          & 76.2  & -0.4     \\
                                               &                        & JMP~\cite{DBLP:journals/corr/abs-2005-08931}           & 76.2  & -0.4     \\
                                               &                        & DMCP~\cite{DBLP:conf/cvpr/GuoWLY20}          & 76.6  & 0.0      \\
                                               &                        & \textbf{DNA}           & \textbf{76.9}  & \textbf{+0.3}     \\ \cmidrule{2-5}                                                
                                               & \multirow{1}{*}{4.1G}  & \textbf{DNA}           & \textbf{77.6}  & \textbf{+1.0}     \\ \cmidrule{2-5} 
                                               
                                               & \multirow{2}{*}{7.8G}  & R50 $\veryshortarrow$ R101 & 78.2  & +1.6        \\
                                               &                        & \textbf{DNA}           & \textbf{78.8}  & \textbf{+2.2} \\ 
\bottomrule
                                               
\end{tabular}

\label{tb:cmp with resnet}
\end{table}

\subsection{Comparison with state-of-the-art methods}
\label{sec:sota}
We compare DNA with state-of-the-art methods on ImageNet, 
including network adaption methods: MorphNet~\cite{DBLP:journals/corr/abs-1711-06798}, Network Adjustment~\cite{DBLP:conf/cvpr/Chen0XLWT20}, BigNAS~\cite{DBLP:conf/eccv/YuJLBKTHSPL20},
network expansion method: compound scaling (EffcientNet)~\cite{DBLP:journals/corr/abs-1905-11946} and network shrinking/pruning methods: MetaPruning~\cite{DBLP:journals/corr/abs-1903-10258}, Joint Multi-dimension Pruning~\cite{DBLP:journals/corr/abs-2005-08931}, FPGM~\cite{DBLP:journals/corr/abs-1811-00250}, and DMCP~\cite{DBLP:conf/cvpr/GuoWLY20}. The results are shown in Table~\ref{tb:cmp with resnet} and Table~\ref{tb:cmp with others}. In the table, $\Delta$ columns indicate the accuracy drop of each method compared with the baseline model in their original paper.
From these tables, DNA outperforms nearly all previous methods in multiple settings.
These results show the superiority of DNA and its potential to unify network pruning, network adaption, and network expansion.
It is worth noting that the state-of-the-art network pruning method DMCP achieves comparable results as our method, but DMCP cannot perform network adjusting under the same or larger computation cost. Moreover, the construction of a larger search space for sub-networks is inevitable in DMCP, which limits its application on networks with a large computation budget and indicates the superiority of DNA.
\begin{table}[t]
\centering
\caption{Comparison with different FLOPs settings in MobileNet v1 with input spatial 128, MobileNet v2.}
\begin{tabular}{l|l|lll}
\toprule
base net                   & FLOPs                       & method        & Top-1 & $\Delta$ \\ \midrule
 \multirow{9}{*}{MBV1(128)}& \multirow{1}{*}{186M} & baseline       & 65.2  & -    \\ \cmidrule{2-5}
                           & \multirow{4}{*}{14M}  & Uniform 0.25$\times$       & 44.6  & -22.8    \\
                           &                             & MorphNet~\cite{DBLP:journals/corr/abs-1711-06798}      & 45.9  & -19.5    \\
                           &                             & NetAdapt~\cite{yang2018netadapt}      & 46.3  & -19.1    \\
                           &                             & \textbf{DNA}           & \textbf{47.2}  & \textbf{-18.3}    \\ \cmidrule{2-5}
                           
                           & \multirow{3}{*}{49M}        & Uniform 0.5$\times$  & 56.3  & -8.9        \\
                           &                             & MorphNet~\cite{DBLP:journals/corr/abs-1711-06798}      & 57.5  & -7.7     \\
                           &                             & \textbf{DNA}           & \textbf{58.9}  & \textbf{-6.3}     \\ \cmidrule{2-5}
                           
                           & \multirow{1}{*}{186M}       & \textbf{DNA}   & \textbf{67.5}  & \textbf{+2.3}        \\ \midrule

\multirow{15}{*}{MBV2}      & \multirow{1}{*}{300M}       & baseline          & 72.4  & -   \\ \cline{2-5}
                           & \multirow{2}{*}{43M}        & Meta~\cite{DBLP:journals/corr/abs-1903-10258}          & 58.3  & -13.7    \\
                           &                             & \textbf{DNA}           & \textbf{59.0}  & \textbf{-12.4}    \\ 
\cline{2-5}
                           & \multirow{3}{*}{59M}        & Uniform 0.35$\times$      & 60.3  & -12.5    \\
                           &                             & DMCP~\cite{DBLP:conf/cvpr/GuoWLY20}          & 62.7  & -10.1    \\
                           &                             & \textbf{DNA}           & \textbf{62.9}  & \textbf{-9.5}    \\ \cmidrule{2-5} 
                           & \multirow{5}{*}{210M}       & Uniform 0.75$\times$ & 70.4  & -2.0     \\
                           &                             & Meta~\cite{DBLP:journals/corr/abs-1903-10258}          & 71.2  & -0.8     \\
                           &                             & JMP~\cite{DBLP:journals/corr/abs-2005-08931}           & 71.6  & -0.8     \\
                           &                             & DMCP~\cite{DBLP:conf/cvpr/GuoWLY20}          & 72.2  & -0.2     \\
                           &                             & \textbf{DNA}           & \textbf{72.4}  & \textbf{-0.0}     \\ \cmidrule{2-5} 
                           & \multirow{2}{*}{300M}       & DMCP~\cite{DBLP:conf/cvpr/GuoWLY20}          & 73.5  & +0.8     \\
                           &                             & \textbf{DNA}           & \textbf{73.5}  & \textbf{+1.1}     \\ \cmidrule{2-5} 
                                                        
                           & \multirow{2}{*}{580M} & Uniform 1.4$\times$      & 75.7  & +3.3     \\
                           &                             & \textbf{DNA}           & \textbf{76.4}  & \textbf{+4.0}     \\  \bottomrule
\end{tabular}

\label{tb:cmp with others}

\end{table}

\begin{table}[h!]
\centering
\caption{Comparison with different FLOPs settings in EfficientNet and MobileNet v3, MBV3(S) and MBV3(L) means MobileNet v3 Small and MobileNet v3 Large settings respectively.}
\begin{tabular}{l|l|lll}
\toprule
\multirow{4}{*}{MBV3(S)}    & \multirow{2}{*}{38M}       & baseline   & 65.4  & -2.0        \\
                           &                             & \textbf{DNA}           & \textbf{66.4}  & \textbf{-1.2}     \\ \cmidrule{2-5}
                           & \multirow{2}{*}{55M}       & baseline  & 67.4 & -     \\
                           &                             & \textbf{DNA}           & \textbf{67.9}  & \textbf{+0.5}     \\ \midrule
\multirow{5}{*}{MBV3(L)}    & \multirow{2}{*}{138M}       & baseline   & 73.3  & -1.9        \\
                           &                             & \textbf{DNA}           & \textbf{74.1}  & \textbf{+0.8}     \\ \cmidrule{2-5}
                           & \multirow{2}{*}{214M}       & baseline  & 75.2 & -     \\
                           &                             & \textbf{DNA}           & \textbf{76.4}  & \textbf{+1.2}     \\ \cmidrule{2-5}
                           & 242M       & BigNAS-S~\cite{DBLP:conf/eccv/YuJLBKTHSPL20}  & 76.5 & -     \\ \midrule
                           
\multirow{4}{*}{Eff-B0}    & \multirow{2}{*}{385M}       & baseline   & 76.2  & -        \\
                           &                             & \textbf{DNA}           & \textbf{76.7}  & \textbf{+0.5}     \\ \cmidrule{2-5}
                           & \multirow{2}{*}{685M}       & B0 $\veryshortarrow$ B1*  & 77.4  & +1.2     \\
                           &                             & \textbf{DNA}           & \textbf{78.2}  & \textbf{+2.0}     \\ \bottomrule

\end{tabular}

\label{tb:cmp with others}

\end{table}

\section{Conclusion}
In this work, we present a new network adaption method with human priors as little as possible.
The proposed differentiable method and the elastic search space make it possible to automatically search the best depth and width setting of an existing network.
Our method can adapt existing networks to a better performance with a given computation budget than most previous methods.
Even the performance of a state-of-the-art network Efficient-B0 can also be further improved.

\newpage
{\small
\bibliographystyle{ieee_fullname}
\bibliography{egbib}
}

\end{document}